\documentclass[10pt,twocolumn,letterpaper]{article}

\usepackage{graphicx}
\usepackage[pagenumbers]{iccv} % To force page numbers, e.g. for an arXiv version
\definecolor{iccvblue}{rgb}{0.21,0.49,0.74}
\usepackage[pagebackref,breaklinks,colorlinks,allcolors=iccvblue]{hyperref}

\title{HisTrackMap: Global Vectorized High-Definition Map Construction via History Map Tracking}
\author{Jing Yang~\thanks{Equal contribution.}\quad\thanks{Work done during an internship at Baidu.}\\
Tongji University\\
\and
Sen Yang~\footnotemark[1]\\
Baidu Inc.\\
\and
Xiao Tan\\
Baidu Inc.\\
\and
Hanli Wang~\thanks{Corresponding authors.}\\
Tongji University\\
}

\begin{document}
\maketitle
\begin{abstract}

As an essential component of autonomous driving systems, high-definition (HD) maps provide rich and precise environmental information for auto-driving scenarios; however, existing methods, which primarily rely on query-based detection frameworks to directly model map elements or implicitly propagate queries over time, often struggle to maintain consistent temporal perception outcomes. 
These inconsistencies pose significant challenges to the stability and reliability of real-world autonomous driving and map data collection systems. 
To address this limitation, we propose a novel end-to-end tracking framework for global map construction by temporally tracking map elements' historical trajectories.
Firstly, instance-level historical rasterization map representation is designed to explicitly store previous perception results, which can control and maintain different global instances' history information in a fine-grained way.
Secondly, we introduce a Map-Trajectory Prior Fusion module within this tracking framework, leveraging historical priors for tracked instances to improve temporal smoothness and continuity.
Thirdly, we propose a global perspective metric to evaluate the quality of temporal geometry construction in HD maps, filling the gap in current metrics for assessing global geometric perception results.
Substantial experiments on the nuScenes and Argoverse2 datasets demonstrate that the proposed method outperforms state-of-the-art (SOTA) methods in both single-frame and temporal metrics.
The project page is available at: \href{https://yj772881654.github.io/HisTrackMap/}{https://yj772881654.github.io/HisTrackMap.}
   % In this paper, We introduce HisTrackMap, a novel end-to-end tracking method for vectorized map construction that utilizes a temporal History map. It temporally constructs and maintains an instance-level history map, which provides geometric prior information for the tracked map elements, 
   % improving the stability of the map construction and progressively generating a global vectorized map.
   % To fully leverage temporal geometric information, 
   % we leverage Map-Trajectory Prior Fusion to the track queries by integrating bird's-eye view (BEV) and perspective view (PV) features.
   % Moreover, we find that the existing single-frame perception paradigm is insufficient for evaluating the stability, continuity, and usability of reconstruction results in real-world autonomous driving scenes, as current State Of The Art (SOTA) models still struggle to produce temporally consistent maps.
   % This paper further establishes a benchmark from a global geometric perspective and introduces G-mAP, a metric designed to comprehensively evaluate model performance in terms of temporal consistency and global geometric continuity.
   %  Experimental results demonstrate that HisTrackMap achieves SOTA performance on the nuScenes and Argoverse2 datasets, excelling in both single-frame and temporal evaluation metrics.

\end{abstract}    
\section{Introduction}

High-definition (HD) maps, which include vectorized map elements such as lane dividers, pedestrian crossings, and road boundaries, play a critical role in the navigation and planning of autonomous driving~\cite{xiong2023neural,xiao2020multimodal,xu2024drivegpt4,prakash2021multi,li2025generative}.
Traditional map construction methods use the SLAM-based method~\cite{shan2020lio,shan2018lego,zhang2014loam} to collect offline map data, followed by extensive post-processing to generate HD maps. 
However, these methods are constrained by significant limitations, including substantial costs, the absence of real-time processing capabilities, and difficulties in accommodating dynamic environments and road updates.
% \begin{figure}[t]
%   \centering
%   \includegraphics[width=\linewidth]{fig/HisTMap-fig1.png}
%   \caption{\textbf{Comparison of Different HD Map construction Paradigms.} (a) Single-Frame, (b) Temporal, (c) HisTrackMap (Ours).
%   The gray square represents detect queries, while the purple square represents temporal propagation queries.
%   }
%   \label{fig:compared_vectorized_map}
%   \vspace{-1.5em}
% \end{figure}

Recent advancements in Perspective View (PV)-to-Bird’s-Eye View (BEV) methods have significantly enhanced vectorized HD map construction such as~\cite{li2022hdmapnet,maptrv2,liao2022maptr,hu2021fiery}. These approaches leverage the DETR-based detection paradigm~\cite{carion2020end} to achieve precise HD map generation. To illustrate the differences across various paradigms. 
%However, the inner jitters of the model prediction and many uncontrollable conditions such as occlusions or low light often lead to inconsistencies in temporal perception, which presents significant challenges in real-world autonomous driving scenarios.
Nevertheless, internal prediction instabilities within the model coupled with uncontrollable environmental factors, such as occlusions or low-light conditions, frequently result in temporal perception inconsistencies, posing substantial challenges for real-world autonomous driving scenarios.
Latent query embedding is used as a stream memory~\cite{wang2024stream,yuan2024streammapnet}, facilitating the propagation of temporal information within a unified latent memory.
Furthermore, MapTracker~\cite{chen2025maptracker} employs a tracking paradigm that further utilizes query propagation and implicit latent memory to associate instance-level map elements across consecutive frames, enhancing temporal consistency.
% To address the inconsistency in single-frame model perception, MapTracker adopts a tracking-based formulation instead of detection, aiming for ultimate temporal consistency. Specifically, it borrows the query propagation paradigm from tracking literature, which explicitly associates road elements across frames. A sequence of memory latents from past frames serves as the memory mechanism to enhance temporal coherence.
In such a framework, a Motion MLP proposed in~\cite{yuan2024streammapnet} is essential for obtaining the next-frame latent memory considering the pose transformation between consecutive frames. Consequently, it employs a transformation loss to guide the Motion MLP, implicitly ensuring the accuracy of the pose transformation.

% While implicit transformations may be insufficient to capture the fine-grained details of HD maps, in practice, a significant amount of information generated during vehicle motion can be reused, minimizing redundant computations. 
%Although a substantial amount of temporal information generated during vehicle navigation can be efficiently reused in practice to reduce redundant computations, implicit encoding may be insufficient to accurately capture the fine-grained details of feature transformations during the motion process.
It is understood that a substantial amount of historical perception map data generated during vehicle navigation could potentially be leveraged as prior knowledge or utilized to reduce computational overhead in overlapping map regions. However, the current approach of implicit encoding through instance queries remains inadequate for precisely recording or preserving past geometric map information.
To fully utilize existing perception results, we propose HisTrackMap, which explicitly maintains instance-level history maps for corresponding map instances under a tracking paradigm. This approach provides prior information for subsequent navigation perception, ensuring smoother and more continuous HD map construction and enhancing the efficiency and accuracy of temporal information propagation.
% To fully leverage existing perception results, as shown in Fig.~\ref{fig:compared_vectorized_map} (c), we propose HisTrackMap, which employs a tracking paradigm and maintains instance-level history maps to enhance the efficiency and accuracy of temporal propagation.
% Fig.~\ref{fig:histmap} shows the history maps at $\mathbf{T}_1$ and $\mathbf{T}_2$, with each color representing a unique map instance. The pink instance is persistently tracked, and its strongly associated history map propagates temporal geometric information in later timestamps.
\begin{figure}[t]
  \centering
  \includegraphics[width=0.45\textwidth]{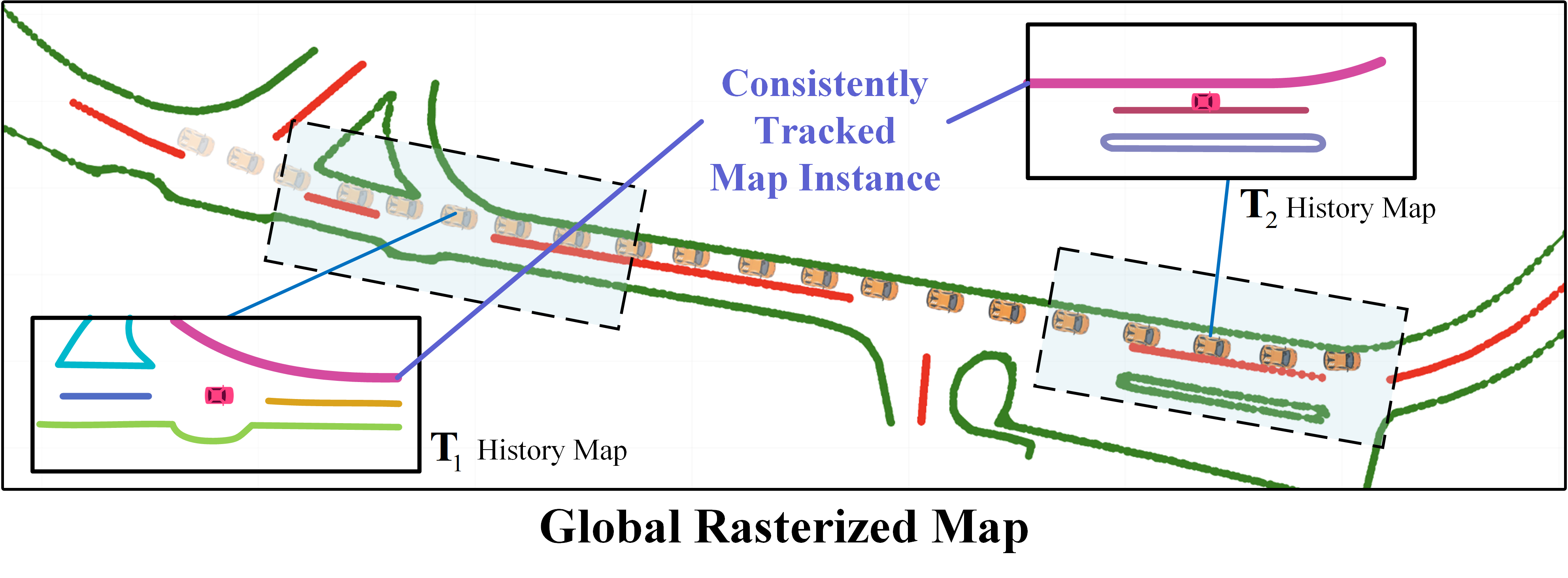}
  \vspace{-1.0em}
  \caption{
% The figure illustrates the progression of a scene’s history map at two distinct timestamps, $\mathbf{T}_1$ and $\mathbf{T}_2$, where each color represents a persistently tracked map instance within the scene.
\textbf{The global rasterized map with history maps.}
Red and green represent divider and boundary categories, respectively. In the history maps, different colors indicate unique tracked map instances. The pink instance is tracked at $\mathbf{T}_1$, consistently propagating temporal geometric information and providing priors for the corresponding track query at $\mathbf{T}_2$.
  }
  % 颜色含义
  \label{fig:histmap}
  \vspace{-1.5em}
\end{figure}
Fig.~\ref{fig:histmap} illustrates the global rasterized map generated from a continuous navigation sequence, along with details of instance tracking.
Furthermore, we propose Map-Trajectory Prior Fusion, which leverages the history map to provide refined prior information for track queries at the subsequent timestamp by integrating position-aligned PV and BEV features.
Our HisTrackMap leverages instance-level history maps to establish one-to-one correspondences between track queries and map trajectories, thereby enhancing global geometric consistency and constructing a temporally consistent vectorized HD map. 

Although traditional Chamfer Distance mean Average Precision (mAP) metrics are widely used for HD map construction, we argue that single-frame metrics are insufficient for evaluating performance in practical autopilot scenarios and map data collection processes.
% and 视觉建图数据采集时也需要
For example, they fail to maintain global constructed map consistency across sequential frames, which is critical to ensure robust long-term perception, reliable decision-making, and efficient data acquisition.
While MapTracker introduced consistency-aware metrics (C-mAP) to penalize inconsistent map elements, they remain indirect methods of evaluation.
To address this issue, we propose a global geometric-aware metric (G-mAP), which directly evaluates the global perceptual quality of the entire scene.
The proposed HisTrackMap consistently outperforms existing methods, demonstrating the effectiveness of utilizing history maps compared to implicit transformations.

In summary, our main contributions are as follows:
\begin{itemize}
\item 
% We propose HisTrackMap, an end-to-end tracking-enhanced framework for vectorized map construction, which leverages temporal information by maintaining instance-level history maps, reducing redundant computations and enhancing efficiency.
An end-to-end tracking-enhanced framework, HisTrackMap, for vectorized HD map construction is proposed, which leverages temporal information by maintaining instance-level history maps, reducing redundant computations, and enhancing efficiency.
\item 
The Map-Trajectory Prior Fusion module is designed to fully utilize the map trajectory information from the history map, combined with instance-level perception features, to provide priors for the corresponding track queries in future frames, thereby optimizing the temporal propagation process.
% We propose the Map-Trajectory Fusion module to fully utilize the map trajectory information from the history map, sampling instance-level perception features to provide priors for the corresponding track queries in future navigation, thereby optimizing the temporal propagation process.
% To fully leverage the historical map, we propose the Map-Trajectory Fusion, which utilizes the trajectory information from the history map to sample the corresponding instances-level perception features, generating priors for track queries in future time steps to optimize the temporal propagation process.
% \item mixing cross-dataset training data to get a stronger instance decdoer
% \item establish a global map evaluation metrics to test.
% mixing cross-dataset training data to get a stronger instance decdoer.
% We integrate HisTracker with two SOTA methods (MapTRv2~\cite{maptrv2} and MapTracker~\cite{chen2025maptracker}), demonstrating remarkable improvements on nuScenes~\cite{caesar2020nuscenes} and Argoverse 2~\cite{wilson2023argoverse} datasets under identical settings. 
\item  
A novel benchmark with a global geometric perspective is introduced to address the limitations of existing evaluation methods in practical applications.
Our HisTrackMap achieves SOTA results in the popular nuScenes~\cite{caesar2020nuscenes} and Argoverse 2~\cite{wilson2023argoverse} datasets.
% To address the shortcomings of global map perspective evaluation, we established a new benchmark to assess the performance of mainstream models.
% Compared to MapTRv2~\cite{maptrv2} and MapTracker~\cite{chen2025maptracker}, 
% HisTrackMap achieves state-of-the-art (SOTA) performance on both the nuScenes~\cite{caesar2020nuscenes} and Argoverse 2~\cite{wilson2023argoverse} datasets in both mAP~\cite{liao2022maptr} and C-mAP~\cite{chen2025maptracker}.
% 此外拟补了一些不足建立了一个新的 benchmark 评估当前模型的表现
% We establish a global geometric map benchmark for comprehensive testing and enhance the instance decoder by integrating cross-dataset training data for greater robustness.

% Furthermore, we integrate HisTracker with two state-of-the-art (SOTA) methods, MapTRv2~\cite{maptrv2} and MapTracker~\cite{chen2025maptracker}, demonstrating remarkable improvements on both the nuScenes~\cite{caesar2020nuscenes} and Argoverse 2~\cite{wilson2023argoverse} datasets under identical settings.
\end{itemize}

\section{Related Work}

% \begin{figure*}[t]
%   \centering
%   \includegraphics[width=0.97\textwidth]{fig/HisTMap-fig2-1.png}
%   % \vspace{-0.5em}
%   \caption{
%  \textbf{The architecture pipeline of the HisTrackMap.}
%  Initialization at timestamp 0 and tracking at $\text{T}$ and $\text{T+1}$ are illustrated.
%  Track queries associate map elements and construct temporal relationships through the history maps under the tracking paradigm.
%  % The pink region will be described in detail later.
%   }
%   % 当前 decoder 输出的 query 里 阈值>pos threshold (positive)/track threshold ，下一帧初始化时作为 tracked query 
%   \label{fig:HisTrackMap}
%   \vspace{-1em}
% \end{figure*}
% T=0 到 T=T 省略号
% pos 角标
% rasterized 可以省略
% history map 一开始的传递
% query 输入输出 track query 和 detect query 区分

 % t=0/T/T+1
 % 第 0 帧的输出 乱一点
  % tracked /positive

\textbf{Map Perception with Single Frame.}
Recently, with advances in PV-to-BEV methods~\cite{yang2023bevformer,li2024bevformer,philion2020lift,liu2022petr,liu2023petrv2,liu2023vectormapnet}, the HD map construction has predominantly been formulated as a detection task using surround-view images from vehicle-mounted cameras in a single-frame setting.
% VectorMapNet~\cite{liu2023vectormapnet} is the first end-to-end HD map construction model using transformers.
MapTR~\cite{liao2022maptr,maptrv2} proposes a unified approach for map elements, addressing ambiguities and ensuring stable learning.
In representing map elements, PivotNet~\cite{ding2023pivotnet} adopts a pivot-based representation, while GeMap~\cite{zhang2025online} models geometric features through Euclidean shapes. 
MGMapNet~\cite{yang2024mgmapnet} leverages multi-granularity representations to model map instances, integrating instance-level and point-level queries.

\textbf{Map Perception with Priors.} 
Some of the latest methods have also focused on integrating rasterized results and standard (SD) maps to provide priors for HD map construction.
SMERF~\cite{smerf} integrates SD maps for online map prediction, with PMapNet~\cite{jiang2024p} enhancing performance by focusing on relevant SD Map skeletons. TopoSD~\cite{yang2024toposd} encodes SD map elements into neural representations and instance tokens, using them as priors for lane geometry and topology decoding.
Mask2Map~\cite{choi2024mask2map} uses a two-stage framework, generating a rasterized map via a segmentation network and refining it into a vectorized map with a mask-driven network.
HRMapNet~\cite{zhang2025enhancing} leverages historical rasterized maps at the city scale and vehicle localization to improve online vectorized map perception.
% smerf toposd pmapnet 

\textbf{Map Perception with Temporal Framework.}
Temporal information is crucial, particularly in complex scenarios with long distances and occlusions, and has been a consistent focus in 3D object detection~\cite{huang2022bevdet4d,wang2023exploring,han2024exploring,lin2023sparse4d}.
StreamMapNet~\cite{yuan2024streammapnet} is the first temporal map construction framework. It employs multi-point attention, enabling the construction of large-scale local HD map with high stability.
SQD-MapNet~\cite{wang2024stream} introduces the Stream Query Denoising (SQD) strategy for temporal modeling.
GlobalMapNet~\cite{shi2024globalmapnet} attempts to perform evaluations from a global perspective and proposes Map NMS to obtain a clean global map.
MapTracker~\cite{chen2025maptracker} presents a vector HD-mapping algorithm that formulates the mapping as a tracking task and leverages latent memory to ensure temporally consistent constructions.

% \textbf{Evaluation Metrics.}  
% MapVR~\cite{zhang2024online} and MGMap~\cite{liu2024mgmap} evaluate rasterized results using IoU (Intersection over Union) metric. 
% In contrast, MapTR~\cite{liao2022maptr,maptrv2} emphasizes point-to-point correspondence and employs Chamfer Distance to calculate average precision.
% However, these metrics are primarily effective for evaluating instantaneous performance.
% Since map perception in practice requires evaluation over continuous sequences, these metrics struggle to measure temporal consistency effectively.
% To address this limitation, MapTracker~\cite{chen2025maptracker} introduces the consistency-aware metric (C-mAP), which aims to penalize mismatches in perception results based on the alignment of corresponding instances across frames.

\section{HisTrackMap}

%------------------------------------------------------------------------
\subsection{Overview Architecture}
% Fig.~\ref{fig:HisTrackMap} illustrates the framework of the proposed HisTrackMap, built on a tracking paradigm~\cite{chen2025maptracker,zeng2022motr}.
The encoder extracts perception features (BEV feature $\mathbf{F}_{bev}$ and PV feature $\mathbf{F}_{pv}$) from surrounding-view images to localize spatial structures.
The input detection query $\mathbf{Q} \in \mathbb{R}^{N_q \times C}$ is updated through MapDecoder to produce the coordinates $\mathbf{P} \in \mathbb{R}^{N_q \times N_p \times 2}$, categories $\mathbf{C} \in \mathbb{R}^{N_q \times 3}$ (pedestrian, boundry, divider) and scores $\mathbf{S} \in \mathbb{R}^{N_q}$. Here, $N_q$ denotes the number of detected queries, $N_p$ represents the number of points of a map instance, and $C$ is the feature dimension.

After the decoder, \textcolor{black}{
the outputted detected queries and propagated track queries are filtered to be positive using a detection threshold $\tau_{det}$ and a tracking threshold $\tau_{track}$. These positive $N_{track}$ queries $\mathbf{Q}_{track}$ are propagated to the next frame as new track queries.
Queries with confidence below the thresholds are treated as disappeared instances and discarded as negative queries.
In summary, map instances will be categorized as newly appeared, consistently tracked, or disappeared, and the corresponding history maps will be initialized, updated with the rasterized map, or removed accordingly.
}
Furthermore, the Map-Trajectory Prior Fusion utilizes the history map to sample corresponding positions on perception features extracted by the encoder, establishing a strong geometric prior for initializing track queries in the new frame and explicitly optimizing the propagation of track queries across temporal sequences.

Section \ref{Sec32} introduces the instance-level history maps for maintaining and updating historical map trajectories.
Subsequently, Section \ref{Sec33} presents the Map-Trajectory Prior Fusion, which provides prior information to enrich track queries.
Finally, Section \ref{Sec34} proposes the metrics to address the current insufficiencies in evaluating stability and continuity in real-world autonomous driving scenarios.

\subsection{Instance-Level History Maps }
\label{Sec32}
% / tracking diagram

We first introduce instance-level maps used for storing historical information.
The history map is maintained for the tracked instances generated during online prediction to store their trajectory information.
During the propagation process, each track query corresponds to a unique map instance and history map, thereby establishing a strong one-to-one relationship.

% \textcolor{red}{
% We use $N_q$ queries for prediction.
% First, the input query $\mathbf{Q}$ obtains the coordinates $\mathbf{P}$, category $\mathbf{C}$, and scores $\mathbf{S}$ for the current frame through $\operatorname{MapDecoder}$:
% \begin{equation}
% \mathbf{P},\mathbf{C},\mathbf{S}=\operatorname{MapDecoder}(\mathbf{Q}).
% \end{equation}}

% \textcolor{red}{
% At the end of each frame, we apply the rasterization method used in MapVR~\cite{zhang2024online} to convert the vectorized map into a rasterized format for each instance.}
% This allows for obtaining a history map with pose transformation when the next frame begins. 
% $\mathcal{M}$ is a collection of history maps corresponding to a series of track queries, defined as $\mathcal{M}=\{\mathbf{M}_1,\mathbf{M}_2 ... \mathbf{M}_{N_{track}}\}$.
% Each $\mathbf{M}_{j}\in \mathbb{R}^{H\times W}$ represents a history map for the $j$-th tracked instance in the current scene, where $H$ and $W$ denote the dimensions of the BEV feature.
% $N_{track}$ represents the total number of tracked instances in the current scene.

% For the map of the $i-th$ instance in the $t$ timestep, when $k$ instances appear in the scene for the first time, specifically at the initial point of tracking, it utilizes an empty history map:
% \begin{equation}
% \begin{aligned}
%     \mathbf{M}_i^{t}&=\operatorname{Zeros}(\mathbf{Q}_i),\\
%     \mathcal{M}^{t} &= \mathcal{M}^{t-1} \cup \{\mathbf{M}_i^t, \mathbf{M}_{i+1}^t,..., \mathbf{M}_{i+k}^t\},
% \end{aligned}
% \end{equation}

%i 和 j 的区分
\textcolor{black}{
At the beginning of frame $t$, $\mathcal{M}$ is a collection of history maps corresponding to a series of track queries, defined as $\mathcal{M}^t=\{\mathbf{M}_i^t \mid i = 1, 2, \dots, N_{track}^{t-1}\}$.
Here, each $\mathbf{M}_{i}^{t} \in \mathbb{R}^{H \times W}$ indicates the rasterized history map of the $i$-th instance among the $N_{track}^{t-1}$ instances that have been tracked at the previous frame, where $H$ and $W$ are equal to the height and width of the BEV feature, respectively. 
}

Subsequently, the perception results at timestamp $t$ are used to update the history map.
For the $j$-th instance, if it is identified as a newly born instance (i.e., no corresponding tracking index is found, and its confidence score $\mathbf{S}_{j}^t$ exceeds the detect threshold $\tau_{det}$), the rasterization method $\operatorname{Raster}(\cdot)$~\cite{zhang2024online} is employed to transform the vectorized map representation $\mathbf{P}_{j}^t$ into a rasterized format. The rasterized result is scaled by $\mathbf{S}_{j}^t$ to initialize its history map.
If the instance has been tracked (i.e., some tracking index is associated, and $\mathbf{S}_{j}^t$ exceeds the tracking threshold $\tau_{track}$), the corresponding historical map $\mathbf{M}_{i}^{t-1}$ is retrieved based on its tracking index $i$. This history map is then temporally decayed using a decay factor $\lambda$ and updated with the newly rasterized result, yielding the updated history map.
% $\mathbf{M}_{i}^{t}$ as $\max\big(\lambda\cdot\mathbf{M}_i^{t-1}, \operatorname{Raster}(\mathbf{P}_j^t) \cdot \mathbf{S}_j^t\big)$.
% \begin{equation}
% \mathbf{M}_i^t = 
% \begin{cases}
%     \operatorname{Raster}(\mathbf{P}_j^t) \cdot \mathbf{S}_j^t, & \text{if newborn}, \\[6pt]
%     \max\big(\lambda\cdot\mathbf{M}_i^{t-1}, \operatorname{Raster}(\mathbf{P}_j^t) \cdot \mathbf{S}_j^t\big), & \text{if tracked}.
% \end{cases}
% \end{equation}
After completing the update process, the history map $\mathbf{M}_{i}^t$ is finalized. At timestamp $t$, both the total number of tracked instances $N_{track}^t$ and the collection of history maps $\mathcal{M}^{t}$ are updated accordingly.
At the $t + 1$ timestamp, if the confidence score exceeds the threshold $\tau_{track}$, the instance will continue to be tracked. 
At timestamp t+1, $\mathcal{M}^{t}$ is dynamically aligned with the updated ego position to obtain warped history maps $\mathcal{M}_w^{t+1}$, thereby enabling initialization for the subsequent timestamp. 
% as
% \begin{equation}
% \mathcal{M}_w^{t+1}=\operatorname{Warp}(\mathcal{M}^{t},\mathcal{T}^{t+1}),
% \end{equation}
% where $\mathcal{T}^{t+1}$ denotes a standard $4 \times 4$ transformation matrix that characterizes the spatial transformation relationship between the coordinate systems of two consecutive frames, $\operatorname{Warp}(\cdot)$ is the alignment function.

Note that, if the prediction confidence of some track query is below the $\tau_{track}$, it indicates that that instance has disappeared or has not been successfully tracked at the current frame. Consequently, it is necessary to remove the corresponding track query and history map $\mathbf{M}_{r}$ from $\mathcal{M}^{t+1}$, and the total number of tracked instances $N_{track}^{t+1}$ will be decremented accordingly.

% as:
% \begin{equation}
% \mathcal{M}^{t+1}=\mathcal{M}_w^{t+1}\backslash \{\mathbf{M}_r\}.
% \end{equation}
% 分类讨论 等号左边不一样

Each map instance follows the process of initialization, tracking, and history map updating to achieve the binding of historical trajectory information with the track query, enabling temporal propagation.

\subsection{Map-Trajectory Prior Fusion}
\label{Sec33}
% \begin{figure}[t]
%   \centering
%   \includegraphics[width=0.45\textwidth]{fig/HisTMap-fig3-1.png}
%   \caption{
%   \textbf{Architecture of Map-Trajectory Prior Fusion.}
%   }
%   \vspace{-1.5em}
%   \label{fig:MPF}
% \end{figure}
As aforementioned, to address the limitations of implicit temporal propagation in capturing detailed feature transformations during motion, the Map-Trajectory Prior Fusion is designed to integrate historical map trajectory information into track queries, enhancing temporal prior utilization and perceptual consistency.
The Map-Trajectory Prior Fusion first samples instance features from the history map within the PV and BEV feature spaces. Since the number of instance features varies, padding is applied to align them to a uniform length. These instance-level perception features are then individually integrated into the track queries through cross-attention.

Historical information provides both semantic category and spatial trajectory coordinate data. 
To enhance the semantic representation of track queries, we define an initial class embedding $\mathbf{CE}_{init} \in \mathbb{R}^{3 \times C}$, which encodes previous track categories into a class embedding $\mathbf{CE} \in \mathbb{R}^{N_{track} \times C}$.
% The class embedding is fused with the track query to provide category priors as}
% \begin{equation}
% \mathbf{Q}_{track}=\mathbf{Q}_{track}+\mathbf{CE}.
% \end{equation}

The history map is a fine-grained map with temporal decay. 
Therefore, we generate the valid pixel mask $\mathcal{M}_{val}$ by filtering pixels in $\mathcal{M}$ that exceed the map threshold $\tau_{map}$.
% To enhance the spatial awareness of PV features in 3D space, learnable coordinate embeddings $\mathbf{PE}_{coords}\in\mathbb{R}^{h\times w\times C}$ are combined with camera embeddings $\mathbf{PE}_{cam}\in\mathbb{R}^{cams\times C}$ generated via the ego-to-image matrix $\mathbf{E2I}\in\mathbb{R}^{cams\times4\times 4}$, producing the final PV position embedding $\mathbf{PE}_{pv}\in\mathbb{R}^{cams\times h\times w\times C}$ as
% \begin{equation}
% \mathbf{PE}_{pv}=\mathbf{PE}_{cam}+\mathbf{PE}_{coords},
% \end{equation}
% where $h, w$ are the resolutions of the PV features, and $cams$ is the number of cameras.}

Then, we project $\mathcal{M}_{val}$ into the perspective view space using the projection function $\operatorname{Proj}(\cdot)$. In this space, the PV features $\mathbf{F}_{pv}$ are combined with the corresponding positional embedding $\mathbf{PE}_{pv}$ to form enhanced feature representations. Subsequently, the $\operatorname{SampledPV}(\cdot)$ function is applied to sample from these enhanced features, resulting in the final sampled PV features

% $\mathbf{F}_{sampled\_pv}$ as
% \begin{equation}
% \mathbf{F}_{sampled\_pv} = \operatorname{SampledPV}(\operatorname{Proj}(\mathcal{M}_{val}, \mathbf{E2I}), \mathbf{F}_{pv}+\mathbf{PE}_{pv}).
% \end{equation}

Similarly, we perform analogous operations in the Bird's-Eye View space.
To ensure that the sampled BEV features $\mathbf{F}_{sampled\_bev}$ incorporate positional information, we introduce a sinusoidal position embedding $\mathbf{PE}_{bev} \in \mathbb{R}^{H \times W \times C}$.
Then, we utilize $\mathcal{M}_{val}$ to sample BEV feature $\mathbf{F}_{bev}$ through the $\operatorname{SampledBEV}(\cdot)$ function.

% as
% \begin{equation}
% \mathbf{F}_{sampled\_bev} = \operatorname{SampledBEV}(\mathcal{M}_{val}, \mathbf{F}_{bev}+\mathbf{PE}_{bev}).
% \end{equation}

Next, the PV and BEV features corresponding to each instance, including the historical map trajectories, have been obtained. The track queries, along with their associated sampled features $\mathbf{F}_{sampled\_bev}$ and $\mathbf{F}_{sampled\_pv}$, are then utilized within a cross-attention $\operatorname{CA}(\cdot)$ to finalize the initialization of the track queries for the current timestamp.

% \begin{equation}
% \begin{array}{l}
% \mathbf{Q}_{track}=\mathbf{CA}(\mathbf{Q}_{track},\mathbf{F}_{sampled\_pv}),\\[6pt]
% \mathbf{Q}_{track}=\mathbf{CA}(\mathbf{Q}_{track},\mathbf{F}_{sampled\_bev}).\\
% \end{array}
% \end{equation}

Finally, we explicitly leverage historical trajectories to provide precise priors for track queries, avoiding redundant auxiliary supervision and inaccurate temporal transformations in implicit propagation.
Notably, since the valid pixels for each instance vary, we pad the features and apply a padding mask to ensure that each track query focuses on its corresponding positions.

\begin{algorithm}[t]
\caption{Global\_Instance\_Match}
\label{alg:ins_mtc}
\begin{algorithmic}[1]
\REQUIRE Distance cost matrix $\mathbf{CM}$, \\
\hspace{2.5em}average confidence scores $\mathbf{AS}$, \\ 
\hspace{2.5em}distance threshold $\tau_{dis}=\{0.25, 0.5, 0.75,  1.0\}$,\\ 
\hspace{2.5em}valid threshold $\tau_{valid}=2$
\ENSURE True positives $\textbf{TP}$, false positives $\textbf{FP}$
\STATE $\textbf{TP} \leftarrow \text{zeros}(\mathbf{CM}.shape[0])$ 
\STATE $\textbf{FP} \leftarrow \text{zeros}(\mathbf{CM}.shape[0])$
\STATE $\text{gt\_covered} \leftarrow \text{zeros}(\mathbf{CM}.shape[1], \text{dtype}=\text{bool})$

\FOR{$i$ in \text{argsort}(-\textbf{AS})}
    \IF{$\mathbf{CM}[i].\text{min}() \leq \tau_{dis}$}
        \FOR{$j$ in $\mathbf{CM}.shape[1]$}
            \IF{$\mathbf{CM}[i][j] \leq \tau_{dis}$ and \text{gt\_covered}[j]==False}
                \STATE $\text{gt\_covered}[j] \leftarrow \text{True}$
                \STATE $\textbf{TP}[i] \leftarrow \textbf{TP}[i] + 1$  // Match multiple ground truth instances
            \ENDIF
        \ENDFOR
    \ELSIF{$\mathbf{CM}[i].\text{min}() > \tau_{dis} + \tau_{valid}$}
        \STATE $\textbf{FP}[i] \leftarrow 0$ // Invalid FP
    \ELSIF{$\mathbf{CM}[i].\text{min}() \leq \tau_{dis} + \tau_{valid}$}
        \STATE $\textbf{FP}[i] \leftarrow 1$ // Valid FP
    \ENDIF
\ENDFOR
\end{algorithmic}
\end{algorithm}

% \subsection{Global Geometric Vector HD Mapping Benchmarks}
\subsection{Global Geometric HD Mapping Evaluation}
\label{Sec34}
Current map construction metrics have the following limitations:
(1) Single-frame mAP~\cite{maptrv2} does not adequately capture the overall quality of a map, as it emphasizes local and instantaneous evaluations while neglecting the assessment of continuity and consistency.
(2) consistency-aware C-mAP~\cite{chen2025maptracker} indirectly addresses consistency but does not explicitly evaluate map completeness.
To address these, we propose a global geometric-aware metric (G-mAP) that evaluates construction quality from a global perspective.

First, the vectorized ground truth elements ${\mathbf{P}^{gt}_{local}}$ from $N_{seq}$ single frames  of a sequence segment are projected into a global coordinate system and rasterized to produce a complete global map containing all map instances ${\mathbf{P}^{gt}_{global}}$ as
\begin{equation}
\mathbf{P}^{gt}_{global}=\operatorname{RasterGlobal}(\{\mathbf{P}^{gt}_{local}\}_{i=1}^{N_{seq}}),
\end{equation}
where $\operatorname{RasterGlobal}(\cdot)$ denotes the process of transforming local coordinates into the global coordinate system and performing rasterization.

To address truncation issues of polygons (e.g., pedestrians) during vehicle motion, it is more effective to evaluate them using IoU metrics~\cite{zhang2024online}. Thus, we adopt the rasterization-based mAP to evaluate closed pedestrian masks with $AP_{polygon}$.
For polyline categories, we use the Merge function~\cite{chen2025maptracker} $\operatorname{Merge}(\cdot)$ to combine the tracking results $\mathbf{P}^{pred}_{local}$ from each frame, obtaining a globally complete instance vector $\mathbf{P}^{pred}_{global}$ as
\begin{equation}
\mathbf{P}^{pred}_{global}=\operatorname{Merge}(\{\mathbf{P}^{pred}_{local}\}_{i=1}^{N_{seq}}).
\end{equation}

To obtain the vector representation of the entire instance, the ground truth mask of the corresponding instance is utilized to extract discrete points via Farthest Point Sampling~\cite{qi2017pointnet++}, enabling the fitting of the complete polyline.
Subsequently, the $AP_{polyline}$ is calculated using the Chamfer Distance metric~\cite{maptrv2}.

% Single-frame polyline results are merged for each tracking instance to generate a globally complete vectorized instance.
% At this stage, we obtain complete predictions and ground truth from a global perspective. 
% For polygons, $\textbf{AP}_{ped}$ is derived from IoU. For polylines, $\textbf{AP}_{div}$ and $\textbf{AP}_{bou}$ are computed using Chamfer Distance to determine FP and TP.

% Finally, by averaging the results across the three categories, we compute G-mAP, achieving a comprehensive evaluation from a global perspective. This approach not only accounts for map categories but also combines the strengths of both rasterized and vectorized representations.
G-mAP leverages the strengths of both rasterized and vectorized representations. It consists of two components: rasterization-based mAP for polygons (e.g., pedestrian) and vectorization-based mAP for polylines (e.g., divider and boundary). By averaging the results across the three categories, G-mAP achieves a comprehensive evaluation from a global geometric perspective.

The $\operatorname{Global\_Instance\_Match}$ function is used to obtain false positives (FP) and true positives (TP) under the Chamfer Distance metric.
We construct a distance cost matrix $\mathbf{CM} \in \mathbb{R}^{N_{pred} \times N_{gt}}$ using $N_{pred}$ predictions and $N_{gt}$ ground truths,  which represents the average distance between the predicted point coordinates and the ground truth for each instance. Additionally, we compute the average confidence scores $\mathbf{AS} \in \mathbb{R}^{N_{pred}}$ for each instance and define a distance threshold $\tau_{dis}$ to evaluate the vectorization-based mAP across different thresholds.
It is worth noting that we have optimized the evaluation algorithm due to the unique characteristics of the global perspective. 
As shown in Algorithm~\ref{alg:ins_mtc}, to prevent overlooking shorter instances that have been accurately identified, our algorithm permits predicted instances to correspond to multiple shorter ground truth instances in the global perspective by incrementing the true positive (TP) count.
% we consider that the merged instance may correctly match multiple ground truth instances in the global view.
% 我们的算法允许一个预测的全局实例匹配多个较短的实例在全局视角下，因为如果不这样会造成短实例被漏检但实际已经被感知到 具体（怎么做） TP=TP+1
% 为了避免短实例被漏检但实际已经被感知到的情况，我们的算法通过TP=TP+1，允许一个预测的全局单独实例在全局视角下匹配多个较短的实例。
% 如算法~\ref{alg:ins_mtc}所示，为了避免遗漏已被准确识别的较短实例，我们的算法通过增加真阳性（TP）计数，允许单个预测的全局实例与全局视图中的多个较短实例相匹配。
% Additionally, there may be cases where actual instances exist but are not labeled. 
% Additionally, there may be cases where actual instances exist that have already been perceived but are not labeled.}
% 但已经感知到了 此外，可能存在实际实例已经被感知到但未被标注的情况。
% To address this, we use a valid threshold $\tau_{valid}$ to ensure the objectivity of the evaluation.
In addition, there are instances in practice that have been perceived but not labelled. To address this issue, we use a validity threshold $\tau_{valid}$ to constrain the counting of false positives (FP).
% 此外，实际中存在已被感知但未被标注的实例。为了解决这一问题，我们使用有效阈值 $\tau_{valid}$ 限制假阳性（FP）的计数，以确保评估的客观性。

% \begin{algorithm}[t]
% \caption{Global\_Instance\_Match}
% \label{alg:ins_mtc}
% \begin{algorithmic}[1]
% \REQUIRE Distance Cost Matrix $\mathbf{M}$, Confidence Scores $\mathbf{S}$, Distance Threshold $\tau_{dis}=\{0.25, 0.5, 0.75,  1.0\}$, Valid Threshold $\tau_{valid}=0.5$
% \ENSURE True positives $\textbf{TP}$, false positives $\textbf{FP}$
% \STATE $\textbf{TP} \leftarrow \text{zeros}(\mathbf{M}.shape[0])$ 
% \STATE $\textbf{FP} \leftarrow \text{zeros}(\mathbf{M}.shape[0])$
% \STATE $\text{gt\_covered} \leftarrow \text{zeros}(\mathbf{M}.shape[1], \text{dtype}=\text{bool})$

% \FOR{$i$ in \text{argsort}(-\textbf{S})}
%     \IF{$\mathbf{M}[i].\text{min}() \leq \tau_{dis}$}
%         \FOR{$j$ in $\mathbf{M}.shape[1]$}
%             \IF{$\mathbf{M}[i][j] \leq \tau_{dis}$ and \text{gt\_covered}[j]==False}
%                 \STATE $\text{gt\_covered}[j] \leftarrow \text{True}$
%                 \STATE $\textbf{TP}[i] \leftarrow \textbf{TP}[i] + 1$  // Match multiple ground truth instances
%             \ENDIF
%         \ENDFOR
%     \ELSIF{$\mathbf{<}[i].\text{min}() > \tau_{dis} + \tau_{valid}$}
%         \STATE $\textbf{FP}[i] \leftarrow 0$ // Invalid FP
%     \ELSIF{$\mathbf{M}[i].\text{min}() \leq \tau_{dis} + \tau_{valid}$}
%         \STATE $\textbf{FP}[i] \leftarrow 1$ // Valid FP
%     \ENDIF
% \ENDFOR
% \end{algorithmic}
% \end{algorithm}

\begin{table*}[h]
\begin{center}
\begin{tabular}{l|c|ccc|ccc|c}
\hline
Method & Epoch &$AP_{ped}$ &$AP_{div}$ &$AP_{bou}$ &mAP &C-mAP &G-mAP &FPS\\
\hline\hline
MapTRv2~\cite{maptrv2}   \hfill\graytext{[IJCV2024]}  &  24  &59.8 &62.4 &62.4 &61.4&41.7&33.9 &14.1\\
StreamMapNet~\cite{wang2024stream}  
\hfill\graytext{[WACV2024]}&24 &61.9&66.3&62.1 &63.4&38.4&34.3 &13.1\\
HRMapNet~\cite{zhang2025enhancing} \hfill\graytext{[ECCV2024]} &24 &65.8 &67.4 &68.5 &67.3 &49.2 &39.7 &10.3\\
MGMap~\cite{liu2024mgmap} \hfill\graytext{[CVPR2024]} &24 &61.8 &65.0 &67.5 &64.8 &43.5 &39.1 &13.4\\
MGMapNet~\cite{yang2024mgmapnet} \hfill\graytext{[ICLR2025]} &24 &64.7&66.1&69.4&66.8 &45.4&37.5 &11.7\\
% Mask2Map~\cite{choi2024mask2map} \hfill\graytext{[ECCV2024]} &24 &70.6&71.3&\textbf{72.9} &71.6&55.8&44.5 &9.2\\
% MapTracker~\cite{chen2025maptracker}{$^\dag$} \hfill\graytext{[ECCV2024]} & 24 &75.3&69.2&71.2 &71.9&63.4&47.3 &10.9\\
% \textbf{HisTrackMap (Ours)} &  24 &&& &73.3 &64.5&48.7\\
% \textbf{HisTrackMap (Ours)} &24 &\textbf{76.9}&\textbf{72.7}&{71.9} &\textbf{73.8}&\textbf{64.7}&\textbf{48.5} &10.3\\
% \hline
% MapTRv2~\cite{maptrv2} \hfill\graytext{[IJCV2024]} &110  &68.1 &68.3 &69.7 &68.7 && &14.1 \\
HRMapNet~\cite{zhang2025enhancing} \hfill\graytext{[ECCV2024]} &110 &72.0 &72.9 &75.8 &73.6 &61.4 &47.9&10.3\\
% MGMap~\cite{liu2024mgmap} \hfill\graytext{[CVPR2024]} &110 &64.4 &67.6 &67.7 &66.5&& &13.4\\
Mask2Map~\cite{choi2024mask2map} \hfill\graytext{[ECCV2024]} &110 &73.6 &73.1 &\textbf{77.3} &74.6 &60.3 &48.8&9.2\\
% MGMapNet~\cite{yang2024mgmapnet} \hfill\graytext{[ICLR2025]} &110 &71.3 &76.0 &73.1 &73.6&&&11.7\\
% MapTracker~\cite{chen2025maptracker} \hfill\graytext{[ECCV2024]} & 72 &80.0 &74.1 &74.1 &76.1 %(74.9/75.3 reproduced) 
% &69.1 &50.7 &10.9\\
MapTracker~\cite{chen2025maptracker} \hfill\graytext{[ECCV2024]} & 72 &\textbf{80.0} &74.1 &74.1 &76.1 
&\textbf{69.1} &49.4 &10.9\\
% 77.0 80.0 73.7 76.9 68.3
% \textbf{HisTrackMap (Ours)} & 72 &79.8&74.5&75.4 &76.6&65.7&51&10.3\\
% \textbf{HisTrackMap (Ours)} & 72 &79.8&74.5&75.4 &76.6&68.7&51.1&10.3\\
\textbf{HisTrackMap (Ours)} & 72 &79.8&\textbf{74.5}&75.4 &\textbf{76.6}&68.7&\textbf{50.2}&10.3\\
\hline
\end{tabular}
\end{center}
\vspace{-1.0em}
\caption{\textbf{Comparison with SOTA methods on the nuScenes validation set.}
$^\dag$ indicates the version reproduced. 
}
\label{Tab:nusc}
\end{table*}

\section{Experiments}
\subsection{Experimental Settings}

\textbf{Dataset.}
We evaluate HisTrackMap on two popular autonomous driving datasets: nuScenes and Argoverse 2. The nuScenes~\cite{caesar2020nuscenes} dataset contains 1,000 scenes, each spanning 20 seconds, with data from six synchronized RGB cameras and detailed pose information. 
The Argoverse 2 dataset~\cite{wilson2023argoverse} includes 1,000 sequences with high-resolution images from seven ring cameras, two stereo cameras, LiDAR point clouds, and map-aligned 6-DoF pose data.
Experiments were conducted on both the old~\cite{maptrv2} and new~\cite{yuan2024streammapnet} dataset splits for comprehensive evaluation.

\textbf{Metrics.}
Following previous work~\cite{maptrv2,chen2025maptracker,yuan2024streammapnet,liao2022maptr,yang2024mgmapnet}, we adopt mean Average Precision (mAP) as the primary evaluation metric. 
Evaluation thresholds are set at 0.5m, 1.0m, and 1.5m. $AP_{ped}$, $AP_{div}$, and $AP_{bou}$ represent average precision for pedestrians, dividers, and boundaries, respectively.
% Chamfer Distance is used to determine whether predictions match the ground truth.
In addition, we employ consistency-aware metric (C-mAP)~\cite{chen2025maptracker} and further introduce a novel global geometric-aware augmented metric (G-mAP), as detailed in Sec.~\ref{Sec34}.

\textbf{Implementation Details.}
Our framework builds on MapTracker~\cite{chen2025maptracker}, which serves as the primary baseline.
% To optimize the temporal propagation process, 
We removed the Motion MLP and the auxiliary transformation loss used for its supervision while keeping other loss functions consistent with MapTracker. 
% We utilized 8 NVIDIA RTX A100 GPUs to train on the nuScenes~\cite{caesar2020nuscenes} dataset for a total of 72 epochs, divided into three stages of 18, 6, and 48 epochs, respectively. 
% Similarly, we trained on the Argoverse2~\cite{wilson2023argoverse} dataset for a total of 35 epochs, divided into 12, 3, and 20 epochs, to ensure consistency with MapTracker~\cite{chen2025maptracker}.
Training on the nuScenes~\cite{caesar2020nuscenes} dataset was conducted on 8 NVIDIA RTX A100 GPUs for 72 epochs across three stages (18, 6, and 48 epochs). Similarly, we trained on the Argoverse2~\cite{wilson2023argoverse} dataset for 35 epochs (12, 3, and 20 epochs) to align with MapTracker. 
Additionally, to ensure alignment with methods such as MapTR~\cite{maptrv2,yang2024mgmapnet,zhang2025enhancing}, we evaluated a shorter 24-epoch training configuration on both datasets. 
To address inaccuracies in the PV projection caused by the dataset's missing Z-axis coordinates, we introduced a complete Map-Trajectory Prior Fusion in Stage 2 to accelerate model convergence. In Stage 3, we exclusively adopted a single BEV Prior to further enhancing model performance.
The hyperparameters are configured as $N_{q}=100$, $N_{p}=20$, $C=512$, $\tau_{det}=0.4$, $\tau_{track}=0.5$, $\tau_{map}=0.5$, $\lambda=0.95$.

\begin{table*}[h]
\begin{center}
\begin{tabular}{l|c|ccc|ccc}
\hline
Method  &epoch &$AP_{ped}$ &$AP_{div}$ &$AP_{bou}$ &mAP &C-mAP &G-mAP \\
\hline\hline
% MapTRv2~\cite{maptrv2} \hfill\graytext{\makebox[1.2cm]{\hfill[IJCV2024]}}  &6&62.9 &72.1 &67.1 &67.4&& & \\
MapTRv2~\cite{maptrv2} \hfill\graytext{\makebox[1.2cm]{\hfill[IJCV2024]}}  &24&62.9 &72.1 &67.1 &67.4&$-$&$-$ \\
% HIMap~\cite{zhou2024himap}  \hfill\graytext{\makebox[1.2cm]{\hfill[CVPR2024]}} &6 &69.0 &69.5 &70.3 &69.6 &-&-&-\\
 % Mask2Map~\cite{choi2024mask2map} \hfill\graytext{\makebox[1.2cm]{\hfill[ECCV2024]}} &6 &68.1 &72.7 &73.7 &71.5 &&&\\
  % MGMapNet~\cite{yang2024mgmapnet} \hfill\graytext{\makebox[1.2cm]{\hfill[ICLR2025]}} &6 &67.1&74.6&71.7 &71.2&&\\
 % \hline
HRMapNet~\cite{zhang2025enhancing} 
 \hfill\graytext{\makebox[1.2cm]{\hfill[ECCV2024]}}&30&65.1 &71.4 &68.6 &68.3 &$-$&$-$ \\
HIMap~\cite{zhou2024himap} 
 \hfill\graytext{\makebox[1.2cm]{\hfill[CVPR2024]}} &24 &72.4 &72.4 &73.2 &72.7 &$-$&$-$\\
 MGMapNet~\cite{yang2024mgmapnet} \hfill\graytext{\makebox[1.2cm]{\hfill[ICLR2025]}} &24 &71.3 &76.0 &73.1 &73.6 &$-$&$-$\\
 % MapTracker~\cite{chen2025maptracker}{$^\dag$}  \hfill\graytext{\makebox[1.2cm]{\hfill[ECCV2024]}} &24 &75.0&78.7&73.1&75.6&63.1&47.4\\
 % \textbf{HisTrackMap (Ours)} &24 &\textbf{77.4}&\textbf{78.9}&\textbf{74.1} &\textbf{76.8}&\textbf{66.6}&\textbf{48.2}\\
 % \hline
 MapTracker~\cite{chen2025maptracker} \hfill\graytext{\makebox[1.2cm]{\hfill[ECCV2024]}} &35 &76.9&79.9&73.6&76.8&68.3 &48.0\\
\textbf{HisTrackMap (Ours)} &35 &\textbf{78.8}&\textbf{80.1}&\textbf{74.0} &\textbf{77.7}&\textbf{69.5}&\textbf{48.6}\\
\hline
\end{tabular}
\end{center}
\vspace{-1.0em}
\caption{\textbf{Comparison with SOTA methods on the Argoverse2 validation set.} The '$-$' symbol indicates that no results are provided.}
% The C-mAP and G-mAP metrics are calculated using the weights provided in the open-source code, the MapTracker~\cite{chen2025maptracker} tool, and our own tools.}
\label{Tab:av2}
\vspace{-0.6em}
\end{table*}

\subsection{Comparision with SOTA Method}
\textbf{Comparison on nuScenes.}
As shown in Table~\ref{Tab:nusc}, we compare HisTrackMap with state-of-the-art methods. 
HisTrackMap demonstrates superior performance at 24 epochs, achieving 73.8 mAP, outperforming MapTracker, which achieves 71.9 mAP, by +1.9 mAP.
In addition, HisTrackMap achieves 64.7 C-mAP and 48.5 G-mAP, outperforming MapTracker by +1.3 C-mAP and +1.2 G-mAP, respectively.
In the 72-epoch experiment, HisTrackMap achieved 76.6 mAP, 68.7 C-mAP, and 50.2 G-mAP respectively.
Furthermore, recent methods like Mask2Map and MGMap use single-frame frameworks with limited temporal modeling, while HisTrackMap outperforms them in all metrics.
Notably, by removing transformation loss supervision of Motion MLP, HisTrackMap achieves about 20\% faster training than MapTracker.
During inference, HisTrackMap runs at 10.3 FPS, slightly slower than MapTracker's 10.9 FPS due to feature sampling overhead, which could be optimized with parallel acceleration in practical applications.
In summary, HisTrackMap improves performance and remains practical for real-time autonomous driving applications.

\textbf{Comparison on Argoverse 2.}
The results on the Argoverse 2 dataset, as presented in Table~\ref{Tab:av2}, further validate the effectiveness of HisTrackMap. In the 24-epoch experiment, HisTrackMap achieves a notable mAP improvement of +8.5 over HRMapNet and +9.4 over MapTRv2. Additionally, HisTrackMap surpasses MapTracker with +1.2 mAP, +3.5 C-mAP, and +0.8 G-mAP at 24 epochs, and +0.9 mAP, +1.2 C-mAP, and +0.6 G-mAP at 35 epochs.
Note that due to the different sampling intervals of HisTrackMap and MapTracker on the Argoverse 2 dataset compared to the other methods, we can only evaluate the mAP metrics.
% Experimental results across multiple datasets indicate that our method outperforms other models in single-frame perception, temporal consistency perception, and global reconstruction from a holistic perspective.
\begin{table}[h]
\centering
\label{tab:comparisons}
\resizebox{1\columnwidth}{!}{
\begin{tabular}{c|l|c|c|c}
\toprule
\textbf{Dataset}      & \textbf{Method}    & \textbf{mAP} & \textbf{C-mAP} &\textbf{G-mAP}\\
\midrule
\multirow{2}{*}{nuScenes} 
    % & StMapNet  &33.9&&\\
    & MapTracker  &40.3	&32.5 &28.0 \\
    % & HisTrackMap  &42.2 & \\
    & HisTrackMap  &41.5 &32.8 &28.5 \\
\midrule
\multirow{2}{*}{Argoverse2} 
   % & StMapNet  &58.1 &&\\
    % & MapTracker  &71.3	&63.1 \\
    % & HisTrackMap  &72.5 &30\\
    % reprocude 24epoch
    % & MapTracker  &69.4	&55.4 \\
    % & HisTrackMap  &69.8 &57.7 \\

    % & MapTracker  &71.3	&63.1 &49.3\\ %paper
    % & MapTracker  &65.6/70.3&51.7/61.3 &48.3\\ %reproduce
    % & HisTrackMap  &67/71.4 &51.4/62.7&49.2 \\
    & MapTracker  &70.3&61.3 &48.3\\ %reproduce
    & HisTrackMap  &71.4 &62.7&49.2 \\
\midrule
\multirow{2}{*}{\shortstack{Unified \\ nuScenes}} 
    & MapTracker  &43.5&34.0&29.9\\
    & HisTrackMap  &44.5&35.1&31.2\\
\bottomrule
\end{tabular}
}
\caption{\textbf{Comparisons on non-overlapping datasets.}}
\label{Tab:newsplit}
\vspace{-1.5em}
\end{table}

\textbf{Comparison on non-overlapping datasets.}
The nuScenes and Argoverse 2 datasets exhibit geographical overlaps~\cite{lilja2024localization}. StreamMapNet~\cite{wang2024stream} proposes a non-overlapping dataset split for them.
The experimental results are shown in Table~\ref{Tab:newsplit}. Our method surpasses MapTracker, achieving improvements of +1.2 mAP, +0.3 C-mAP, and +0.5 G-mAP on nuScenes, and +1.1 mAP, +1.4 C-mAP, and +0.9 G-mAP on Argoverse 2.
We conducted cross-dataset experiments to demonstrate further the superior generalization of HisTrackMap. 
The encoder and decoder were initialized with weights from nuScenes and Argoverse 2 respectively, and finetuned for 12 epochs on the nuScenes newsplit before testing.
HisTrackMap achieved 44.5 mAP, 35.1 C-mAP, and 31.2 G-mAP, demonstrating superior generalization performance over MapTracker.

\subsection{Quantitative Evaluations}
\begin{figure*}[t]
  \centering
  \includegraphics[width=0.96\textwidth]{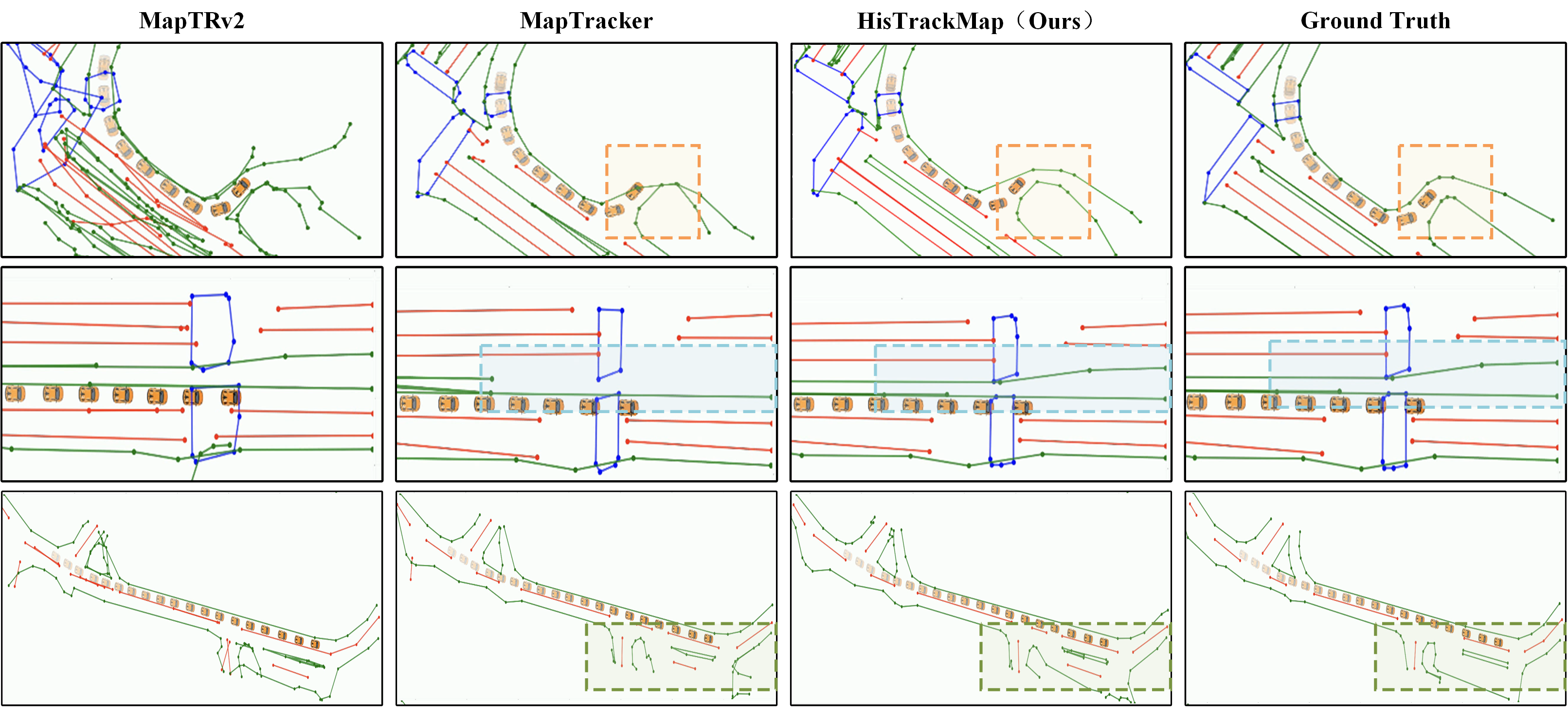}
  \caption{
 \textbf{Qualitative visualization on nuScenes val set.} 
 % The regions marked with rectangles emphasize the superior results generated by our proposed model.
 % Best viewed in color.
  }
  \vspace{-1.0em}
  \label{fig:quantitative_evaluations}
\end{figure*}
Fig.~\ref{fig:quantitative_evaluations} presents a qualitative comparison between HisTrackMap and other methods on nuScenes dataset. 
We utilized the integration~\cite{chen2025maptracker} of per-frame vectorized HD maps into a global vectorized HD map for better visualization.
The rectangular regions highlight instances where our proposed model exhibits superior perception performance.
The orange markings indicate that HisTrackMap accurately identified the two separate boundaries at the turning, whereas MapTracker failed to distinguish them.
The blue markings highlight that HisTrackMap achieved continuous and consistent tracking through a complex intersection, whereas MapTracker lost the polyline.
The green markings indicate that HisTrackMap fully detected the boundary, while MapTracker captured only the first half.
In this green example, all models detected a divider that was not actually annotated, emphasizing the necessity of optimizing false positive (FP) counting in G-mAP.
Furthermore, although single-frame MapTRv2 can partially perceive results, the integrated results lack sufficient stability and fail to maintain consistent perception.
In general, HisTrackMap delivers more accurate and cleaner results, demonstrating superior performance in terms of overall quality and temporal consistency.

% MapTracker produces more accurate and cleaner results, demonstrating superior overall quality and temporal consistency. 
% In scenarios where vehicles turn or do not move straightforwardly, StreamMapNet and MapTRv2 may generate unstable results, leading to fragmented and noisy merged outputs. This is primarily because detection-based formulations struggle to maintain temporally coherent reconstructions under complex vehicle motions.

\subsection{Ablation and Robustness Studies}
The proposed Map-Trajectory Prior Fusion consists of three main components: Class Embedding, PV Prior, and BEV Prior. 
To evaluate the contributions of each component, we conducted an ablation study on the nuScenes dataset. 
The experiment without Map-Trajectory Prior Fusion is used as the baseline.
There are several observations from the results.
First, introducing Class Embedding improved mAP by 0.7, demonstrating the effectiveness of semantic category priors. 
Second, incorporating PV Prior and BEV Prior based on Class Embedding further improves mAP from 72.3 to 72.9 and 73.3, respectively.
This reveals that it is beneficial to utilize geometric trajectory information to enhance track queries.
Third, with the complete Map-Trajectory Prior Fusion, the model achieved the highest 73.8 mAP,
highlighting the significance of history map based trajectory priors in effectively leveraging temporal-spatial information to enhance perception performance.
The update of the history map primarily relies on the vehicle's motion pose parameters. 
To evaluate the robustness of HisTrackMap to localization errors, we conducted additional experiments on the nuScenes dataset.
Random noise was introduced to the translation and rotation of the extrinsic matrix to disrupt the history map update.
The results show that the model achieves 73.0 mAP under noise levels of 0.2 m and 0.02 rad and 73.4 mAP under real-world noise levels (0.1 m, 0.01 rad), maintaining a clear advantage.
The experimental results continue to outperform most previous single-frame models, demonstrating the practical effectiveness of HisTrackMap.

\section{Conclusion}
In this work, we introduce a novel method for end-to-end vectorized HD map construction via tracking history maps, enabling more robust and efficient temporal association modeling. Specifically, the history map is systematically constructed and updated based on past perception results, thereby minimizing redundant computations. 
We introduce the Map-Trajectory Prior Fusion method, which integrates historical map data with current perception features to improve the precision of frame-to-frame transformations. Additionally, we propose a global geometric HD mapping evaluation framework that focuses on assessing overall map construction quality. This evaluation is vital for both autonomous driving systems and map data collection processes.
In the future, it is desired to extend the current work by exploring a more robust, reliable, and adaptive perception system considering the existence of vehicle localization errors.

{
    \small
    \bibliographystyle{IEEEtranN}
    \bibliography{main}
}

\end{document}